\documentclass[fleqn,10pt]{wlscirep}
\usepackage[utf8]{inputenc}
\usepackage[T1]{fontenc}
\usepackage{graphicx}
\usepackage{multirow}
\usepackage{siunitx}
\title{Combining Residual U-Net and Data Augmentation for Dense Temporal Segmentation of Spike Wave Discharges in Single-Channel EEG}

\author[1,*,+]{Saurav Sengupta}
\author[2,+]{Scott Kilianski}
\author[1]{Suchetha Sharma}
\author[2]{Sakina Lashkeri}
\author[2]{Ashley McHugh}
\author[2]{Mark Beenhakker}
\author[1]{Donald E. Brown}
\affil[1]{University of Virginia, School of Data Science, Charlottesville, 22903, USA}
\affil[2]{University of Virginia, School of Medicine, Charlottesville, 22903, USA}

\affil[*]{ss4yd@virginia.edu}

\affil[+]{these authors contributed equally to this work}

\begin{abstract}
Manual annotation of spike-wave discharges (SWDs), the electrographic hallmark of absence seizures, is labor-intensive for long-term electroencephalography (EEG) monitoring studies.  While machine learning approaches show promise for automated detection, they often struggle with cross-subject generalization due to high inter-individual variability in seizure morphology and signal characteristics. In this study we compare the performance of 15 machine learning classifiers on our own manually annotated dataset of 961 hours of EEG recordings from C3H/HeJ mice, including 22,637 labeled SWDs and find that a 1D U-Net performs the best. We then improve its performance by employing residual connections and data augmentation strategies combining amplitude scaling, Gaussian noise injection, and signal inversion during training to enhance cross-subject generalization. We also compare our method, named AugUNet1D, to a recently published time- and frequency-based algorithmic approach called "Twin Peaks" and show that AugUNet1D performs better on our dataset. AugUNet1D, pretrained on our manually annotated data or untrained, is made public for other users.
\end{abstract}
\begin{document}

\flushbottom
\maketitle
%
%
\thispagestyle{empty}

\noindent 


\section*{Introduction}
Childhood Absence Epilepsy (CAE) is characterized by prominent electrographic signals and concomitant loss of consciousness \cite{panayiotopoulos2008typical,tenney2013current}. This pediatric form of epilepsy can significantly impact academic performance and social development, as seizures often occur dozens of times a day, causing frequent lapses in attention \cite{caplan2008childhood,loring2014paying}. Understanding the underlying neural mechanisms of CAE will advance the development of more targeted treatments and improve diagnostic precision. Clearly, accurate and efficient seizure detection is a necessary component for such development.

Electroencephalography (EEG) remains essential for detecting absence seizures. EEG recordings of absence seizures reveal a distinctive pattern of generalized 3-Hz spike-wave discharges (SWDs) that appear suddenly across much of the brain nearly simultaneously. The SWD serves as a diagnostic hallmark that distinguishes absence epilepsy from other types of epilepsy and attention disorders \cite{panayiotopoulos2008typical,tenney2013current}. Indeed, as absence seizures often occur without obvious external symptoms, EEG recordings are necessary for proper diagnosis.

Manual evaluation of EEG recordings remains the gold standard for seizure phenotype assessment, largely due to the inadequacy of current automatic detection algorithms.  Machine learning and neural network approaches offer opportunities for automating this process. In fact, while these approaches have existed for decades\cite{jando1993pattern}, innovations over the last several years have culminated in solutions that automatically label events in raw or minimally preprocessed EEG signals \cite{johansen2016epileptiform,acharya2018deep,ullah2018automated}. These approaches extract hierarchical representations from the raw data rather than rely on features extracted from time, frequency, and time-frequency domains \cite{wang2017automatic,pfammatter2019automated,lestari2020epileptic}. However, such approaches have not yet been tested for the specific use case of automatically detecting SWDs.

Hybrid convolutional neural network-long short-term memory (CNN-LSTM) architectures have shown particular promise by combining spatial feature extraction from CNNs with temporal modeling capabilities of LSTMs \cite{xu2020one,cao2025hybrid}. More recently, specialized architectures designed specifically for EEG analysis (e.g., EEGNet\cite{lawhern2018eegnet}), time series classification (e.g., InceptionTime\cite{fawaz2020jonathanweber}), and dense segmentation tasks (e.g., UNet-based models\cite{ronneberger2015u,perslev2019u}) have emerged as promising approaches for automated physiological signal analysis. Additionally, transformer-based architectures adapted from computer vision, such as DETRtime\cite{wolf2022deep} for temporal event detection and SalientSleepNet\cite{jia2021salientsleepnet} for sleep stage classification, have also been applied to EEG time series analysis.

Considering these numerous approaches to automated event detection, here, we compared the performance of 15 machine learning classifiers to identify optimal SWD detection strategies in the C3H/HeJ mouse model of absence epilepsy \cite{frankel2005c3h,beyer2008c3h,ellens2009c3h}. After determining that the 1D residual U-Net performed best, we further improved performance by applying augmentation procedures to the training data. The final result, AugUNet1D, consistently performed well across recordings from 10 mice in our test dataset. We also compared AugUNet1D to a recent algorithmic method using time- and frequency-based features called "Twin Peaks"\cite{iotchev2024twin}. AugUNet1D again showed superior performance, detecting events with features similar to manually labeled SWDs. Overall, our results demonstrate that AugUNet1D is an effective and reliable method for labeling SWDs in EEG from C3H/HeJ mice. Moreover,  we establish a workflow for utilizing an untrained version of AugUNet1D that can be applied to detect a diverse set of electrographic events across different biological systems.

\section*{Materials and Methods}


\begin{figure}[htbp]
\centering
\includegraphics[width=\linewidth]{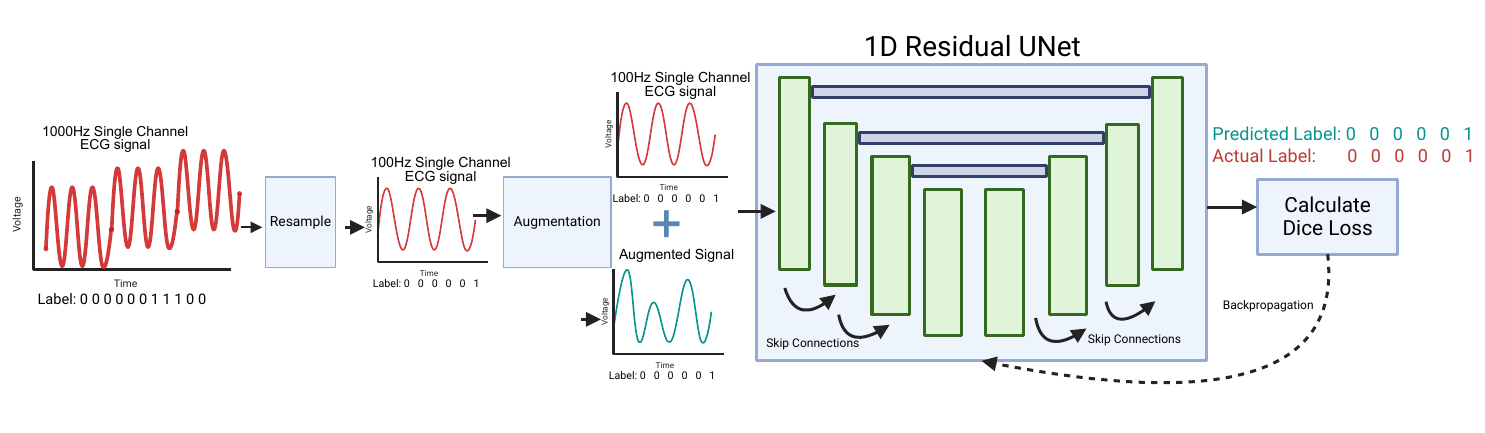}
\caption{Schematic of EEG data preprocessing, augmentation, and the architecture of the AugUNet1D network. We resample all signals to 100Hz, use augmentations like scaling, inversion and Gaussian noise and apply them randomly to the training signal, and then feed them into a residual 1D-UNet that predicts labels at every time point. We minimize Dice Loss using an Adam optimizer.}
\label{fig:fig1}
\end{figure}


\subsection*{Dataset collection}

\subsubsection*{Animals}
The methods described herein are reported following ARRIVE guidelines (https://arriveguidelines.org/). All procedures were approved by the University of Virginia Animal Care and Use Committee (Charlottesville, VA, USA) and performed in accordance with the NIH Guide for the Care and Use of Laboratory Animals. Animals were housed at 23-25\si{\degree C} under an artificial 12-hour light-dark cycle with food and water ad libitum. Male and female C3H/HeJ mice (Strain \#:000659) aged 6-10 weeks were purchased from The Jackson Laboratory and used for the experiments described here. 

\subsubsection*{Electrode Implant Surgery}

Animals were anesthetized with vaporized isoflurane (2-4\% for induction, 1-2\% for maintenance) mixed with pure oxygen and delivered at a rate of 1 liter/min. The scalp was first shaved and then depilatory cream was applied to remove all remaining hair. The scalp was then cleaned with alternating scrubs of betadine and 70\% ethanol three times. After creating a sagittal incision, the underlying fascia was scrubbed away with sterile cotton tipped applicators and an applicator dipped in hydrogen peroxide (3\%) was used to clean the skull and increase the visibility of cranial sutures. Burr holes were made above the secondary motor or primary somatosensory cortex. An additional burr hole was made in the center of the interparietal bone overlying the cerebellum. A custom electrode array, consisting of insulated stainless steel wires (A-M systems part no. 791400) with approximately 500µm insulation stripped from the end soldered to a 6-position header of connector pins (Digikey part no. 1212-1786-ND), was then lowered into the surgical field. The wires were carefully placed under the skull in the burr holes using forceps and set into place with UV-curing glue (SDI Wave). When all wires were positioned and glued, the electrode array was cemented to the remaining exposed skull with dental cement (Metabond, Parkell part no. S380). Skin at the perimeter of the incision was pulled tightly around the cement base and secured with cyanoacrylate adhesive or sutured together at the posterior end of the incision. Ketoprofen (5mg/kg) was administered peri-operatively to reduce pain and inflammation. Animals were given at least seven days to recover prior to EEG recording.

\subsubsection*{Data Acquisition}

EEG signals were first passed through an operational amplifier to buffer small voltage fluctuations before subsequent amplification. These amplifiers were integrated into custom-made PCBs which connected to the header of pins implanted on the head. In some cohorts, signals were then further amplified 10,000 times by a differential amplifier (A-M systems model 3500) and digitally sampled at 1kHz (AD Instruments Powerlab 16/35 or PowerLab C using LabChart) or 400Hz (Grass AURA-64 LTM using TWin software). In all experiments, the cerebellar electrode served as the common reference for other recording channels. Recordings were continuous--24 hours per day--and were stopped, saved, and restarted once per day. All data analyzed in these experiments was recorded from the electrode in the secondary motor cortex. After the final recordings session, animals were deeply anesthetized with a lethal intraperitoneal injection of 60mg/kg Euthasol (pentobarbital sodium and phenytoin sodium) and then, after the loss of pedal reflex was confirmed, transcardially perfused with phosphate-buffered saline followed by 4\% paraformaldehyde. 

\subsubsection*{Manual Labeling of Spike-Wave Discharges}

SWDs were labeled manually by three trained researchers. The following criteria for labeling SWDs were used: (1) the electrographic event must include at least 5 clear rhythmic spike-wave complexes; and (2) the complexes must occur at short, regular intervals separated by at least 50 milliseconds.  Manual labeling was performed in Clampfit (Molecular Devices, San Jose, CA) by placing cursors to bracket the entire SWD (Figure 2). The duration of manual labeling sessions was recorded.

\begin{figure}[h]
\centering
\includegraphics[width=0.7\linewidth]{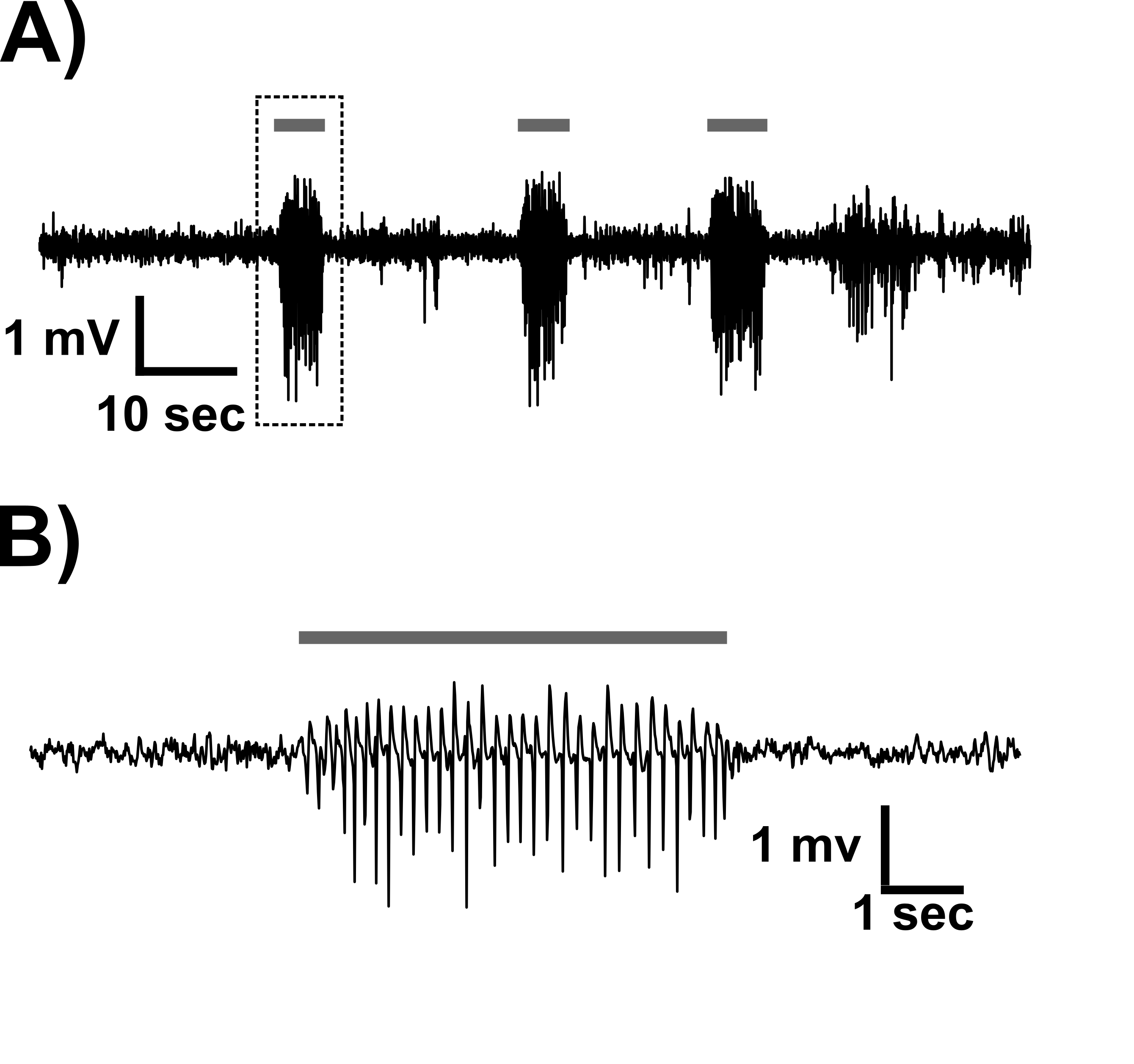}
\caption{Examples of SWDs in raw EEG traces. A) An example EEG trace showing 3 SWDs, each indicated by gray horizontal lines that correspond to manual labels. B) Expanded view of the portion of the trace within the dotted box in A. High amplitude, rhythmic spike-and-wave patterns that define SWD are clearly visible.}
\label{fig:fig2}
\end{figure}

The training data consisted of five days of continuous EEG recording in eight mice, totaling 961.3 hours of recording. On average, it required one hour of manual work to label 8.89 hours of EEG data (Table 1).

\begin{table}[h]
\centering
\begin{tabular}{cccc}
\hline
\textbf{Scorer} & \textbf{Duration to Label (Hours)} & \textbf{Duration of Data (Hours)} & \textbf{Data-To-Labeling Ratio} \\ \hline
Scorer \#1      & 27.3                          & 240.5               & 8.81                                 \\
Scorer \#2      & 38.86                   & 360.4               & 9.27                                \\
Scorer \#3      & 41.93                   & 360.4               & 8.6                                \\
\hline
\textbf{Total}  & 108.09                         & 961.3               & 8.89                      
\end{tabular}
\caption{\label{tab:scorers}Duration (hours) spent scoring the manually labeled dataset}
\end{table}


In total, 22,637 SWDs were manually labeled across five recording days in eight mice (Figure 3A). SWD occurrence was variable across mice, ranging from 1,115 (4.9\%) to 4,570 (20.2\%) SWDs per mouse. The average SWD rate across all mice was 23.55 ± 9.3 per hour (mean ± SD, \textit{N} = 8 mice). The average duration across all SWDs was 5.83 ± 2.37 seconds and varied little between mice (Figure 3B). We also characterized the spectral composition of all manually labeled SWDs (Figure 3C). To this end, the power spectral density of each manually labeled SWD  was calculated, from which the peak frequency was determined (Figure 3C, bottom). The mean peak frequency of all SWDs was 5.72 ± 0.75Hz, again showing little variation across mice (Figure 3D). The observed event rates, duration, and spectral features in our training dataset are very similar to those in previous reports \cite{frankel2005c3h, beyer2008c3h, ellens2009c3h}. Thus, SWD features are generally preserved in C3H/HeJ mice, despite variable SWD occurrence across individuals.

\begin{figure}[htbp]
\centering
\includegraphics[width=0.8\linewidth]{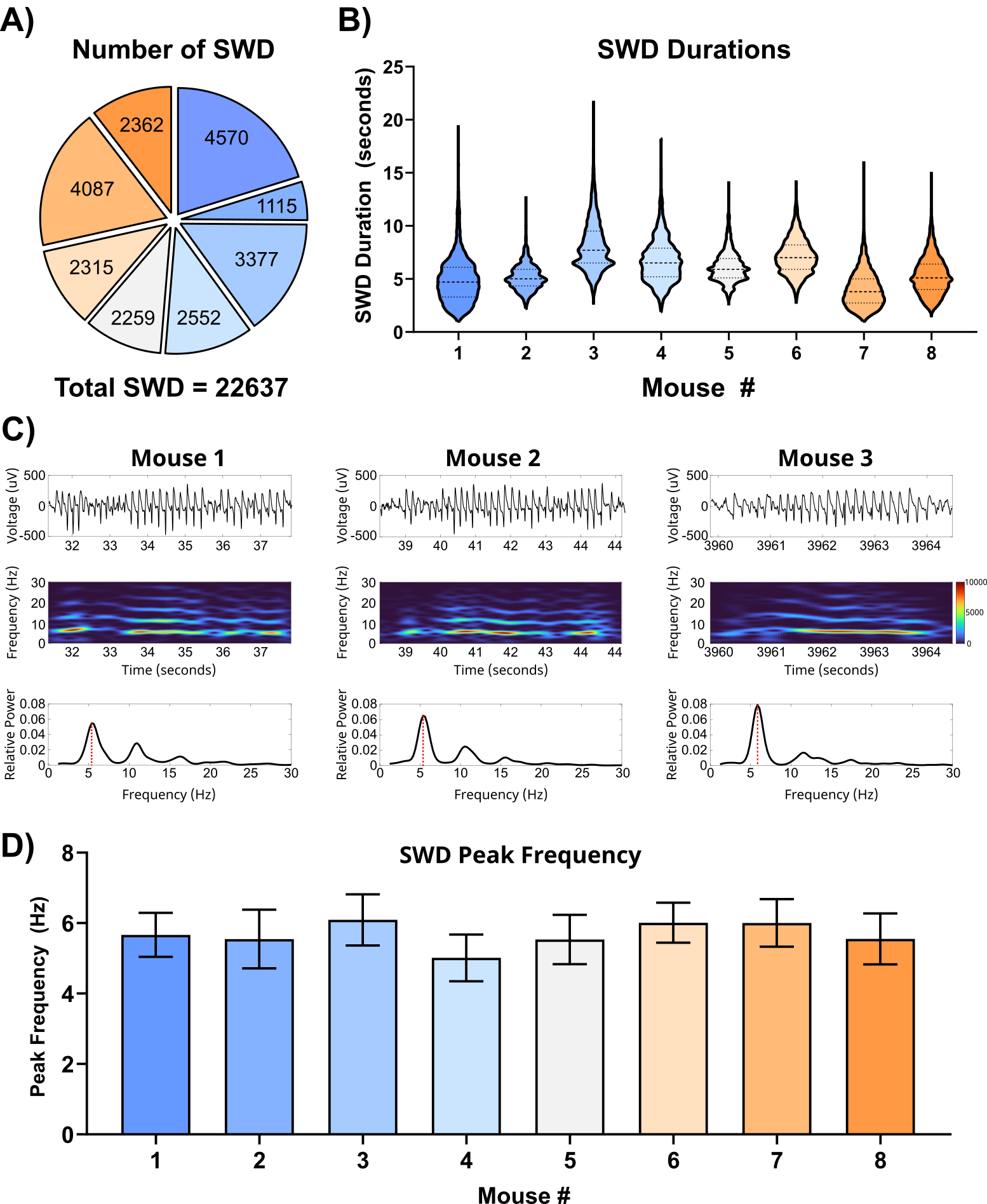}
\caption{Training dataset statistics and spectral features. A) Distributions of seizure durations (in seconds) shown separately for each mouse.  B) Each portion of the whole corresponds to a single mouse. The size of the portion and the number correspond to the number of SWDs in a given mouse. C) Top: Raw EEG data showing example SWDs in three different mice. Middle: Corresponding spectrograms showing power of the signals at different frequencies across time during the events. Bottom: Power spectral density plots showing the average power at different frequencies across the entire events. Vertical red lines indicate the peak frequency of the corresponding event. D) The average (± SD) of the peak frequencies of SWDs in the 8 mice included in the training dataset}
\label{fig:fig3}
\end{figure}

\subsubsection*{Classifying Noise and Sleep in EEG Records}
Because EEG power broadly increases during periods of mechanical noise and during sleep, automated methods often falsely detect events during these periods. To assess how various methods perform during such conditions, noise and sleep epochs must be identified in our recordings. To define noise epochs, first, the mean and standard deviation of EEG amplitudes across the entire recording are computed. Then, traces are segmented into consecutive 5-second blocks. Blocks containing amplitudes 20 standard deviations or more beyond the mean are classified as noise (Figure 4A and B, yellow shading). 

Sleep epochs are defined as periods with elevated power in the 0.1-4Hz frequency band (Figure 4C and 4D, blue shading). Thus, to identify sleep, the instantaneous amplitude (Hilbert envelope) of the 0.1-4Hz-band EEG is first computed as the magnitude of the analytic signal. Then, a distribution of the instantaneous amplitudes is estimated (Figure 4E). This distribution is bimodal: the lower-amplitude peak corresponds to wakefulness. The higher-amplitude peak corresponds to slow wave sleep. The midpoint between the two peaks is used as the lower threshold (red dotted line in Figures 4D and 4E)  to detect the on- and off-set of sleep. Any epoch that does not cross a secondary threshold (the higher-value peak; blue dotted line in Figures 4D and 4E) is not considered sleep. Sleep epochs occurring within 20 seconds of each other are merged.

\begin{figure}[h]
\centering
\includegraphics[width=0.7\linewidth]{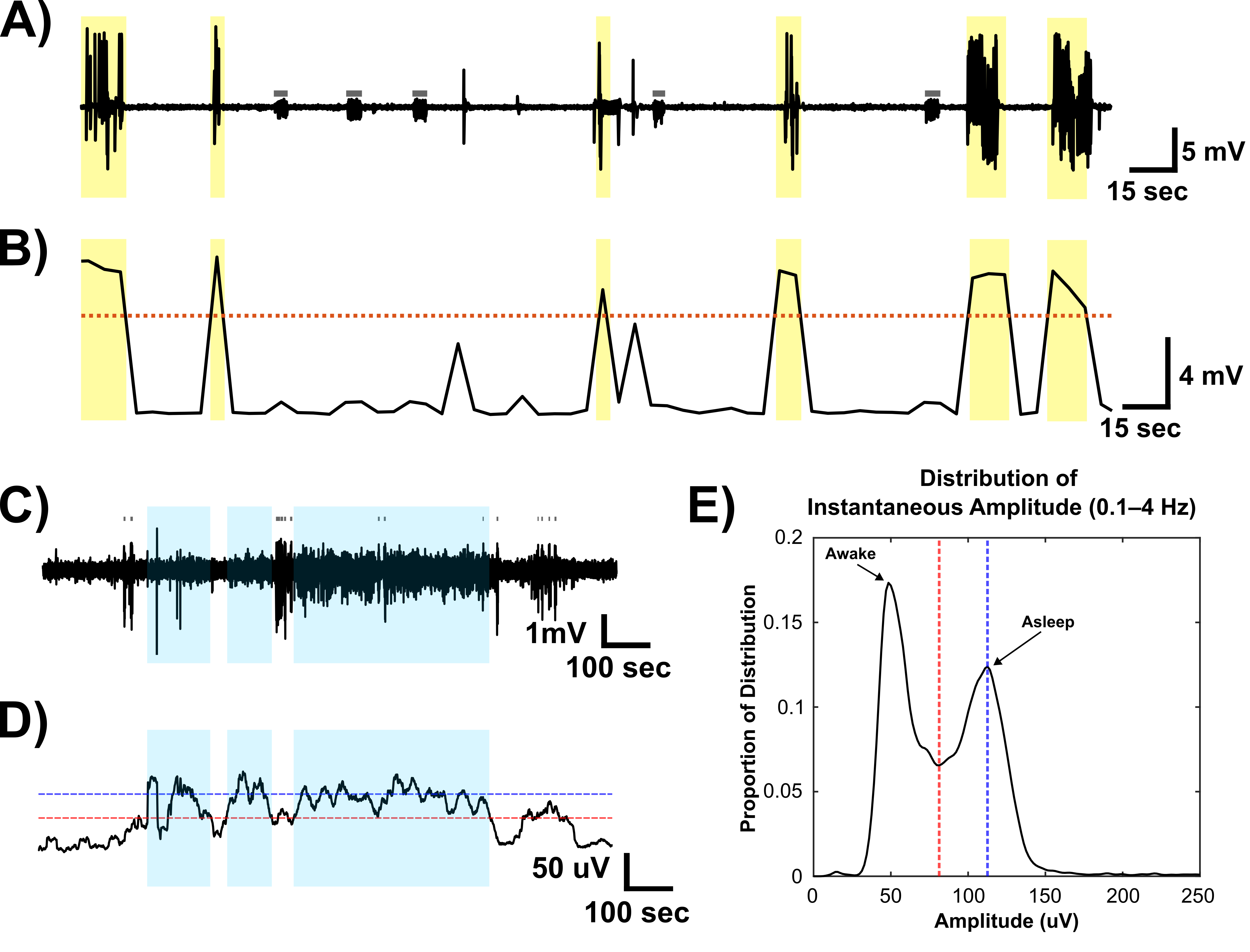}
\caption{Examples of noise and sleep epochs in EEG traces. A) EEG trace showing 6 discrete noise epochs identified by yellow shading. B) Trace shows the maximum absolute EEG amplitude within consecutive 5-second windows. The orange dotted horizontal line is the threshold used to detect noise and represents 20 standard deviations above the overall absolute mean EEG amplitude. C) Representative EEG trace with sleep epochs indicated by blue shading. D) The instantaneous amplitude derived from the Hilbert envelope of the 0.1-4Hz-filtered EEG. The red dotted horizontal line indicates the minimum detection threshold that determines onset and offset of sleep epochs. The trace must also cross the blue dotted horizontal line at least once during an epoch to be classified as sleep. E) Distribution of the amplitude from the Hilbert envelope of the 0.1-4Hz-filtered EEG. This distribution is generated from the entire recording and therefore contains data recorded when the mouse is awake (low amplitude peak) and asleep (high amplitude peak). Horizontal gray bars in A and C indicate manually labeled SWDs.}
\label{fig:fig4}
\end{figure}



 




\subsection*{Dataset pre-processing}
\subsubsection*{Resampling using torchaudio}

To account for variable sampling rates, we used the torchaudio package established by the PyTorch project to resample all input data to a standard sampling rate of 100Hz.
`sinc\_interp\_hann' is a PyTorch function that performs sinc interpolation with Hann windowing for resampling audio or other time-series signals. 


In the context of EEG signal processing for our use case, `sinc\_interp\_hann' is useful for standardizing sampling rates across different recording devices or datasets, ensuring that neural network models receive consistently sampled input data. The associated high-quality interpolation is especially important for EEG analysis because it preserves the precise timing and frequency characteristics of brain signals that are critical for accurate seizure detection or other neurological event classification strategies. 

Therefore, our EEG signal and ground truth labels $x[n]$ are resampled using $
\mathcal{R}_{f_s \rightarrow f_t}: \mathbb{R}^{N_s} \rightarrow \mathbb{R}^{N_t}
$

Where $\mathcal{R}_{f_s \rightarrow f_t}$ is the torchaudio Resample transform that converts an EEG signal from its source sampling rate $f_s$ to its target sampling rate: $f_t$
$$
\tilde{x}(n) = \mathcal{R}_{f_s \rightarrow f_t}(x(n)) = \sum_{k=-\infty}^{\infty} x(k) \cdot h_{\text{sinc}}(n - k \cdot r)
$$

where 
$x[n] \in \mathbb{R}^{N_s}$: Input EEG signal with $N_s$ samples
$\tilde{x}[m] \in \mathbb{R}^{N_t}$: Resampled EEG signal with $N_t$ samples
$f_s$: Source sampling frequency (Hz)
$f_t$: Target sampling frequency (Hz)
$r$: Resampling ratio

\subsection*{Model Building}

For our seizure detection task, we adopted a residual 1D U-Net architecture. The U-Net was originally developed by Ronneberger et al. (2015) \cite{ronneberger2015u} for biomedical image segmentation, where it demonstrated remarkable success in precisely delineating cellular structures and tissue boundaries in microscopy images. The architecture's key innovation is its symmetric encoder-decoder structure with skip connections: the encoder progressively downsamples the input to capture hierarchical spatial features and contextual information, while the decoder upsamples to recover fine-grained spatial details. Crucially, skip connections bridge corresponding encoder and decoder layers, allowing the network to combine high-level semantic features with low-level spatial precision, an essential property for accurate pixel-wise segmentation.

The adaptation of U-Net to temporal segmentation tasks has gained significant traction in recent years across diverse physiological signal processing applications. Perslev et al. (2019) \cite{perslev2019u} pioneered this direction with U-Time, a fully convolutional temporal network for sleep stage classification from EEG data, demonstrating that the U-Net architecture could be successfully applied to sequential physiological signals without requiring recurrent layers. Their work showed that skip connections naturally preserve critical temporal features across multiple time scales, enabling accurate segmentation of sleep stages. Since then, U-Net-based architectures have been successfully deployed for various sequence-to-sequence tasks including seismic phase detection (Zhu \& Beroza, 2019 \cite{zhu2019phasenet}), and more recently, seizure detection from continuous EEG recordings (Wu et al., 2025 \cite{wu2025seizuretransformer}).

The integration of residual connections with U-Net architectures has a well-established history in computer vision, where residual U-Net variants have demonstrated substantial improvements over vanilla U-Net. Zhang et al. (2018) \cite{zhang2018road} introduced ResUNet for road extraction from aerial images, whereas Alom et al. (2018) \cite{alom2018recurrent} proposed the Recurrent Residual U-Net (R2U-Net) for medical image segmentation. Building on these successes, we incorporated residual connections within each encoding and decoding block of our 1D U-Net architecture. These shortcut connections, inspired by ResNet \cite{he2015deep} architectures, allow gradients to flow directly through the network during backpropagation, addressing the vanishing gradient problem and enabling stable training of deeper networks.

Additionally, recognizing that seizure morphologies vary substantially across individuals, we implemented a comprehensive data augmentation strategy during training. Specifically, we randomly applied amplitude scaling (p=0.5) to account for inter-subject variability in EEG amplitude, Gaussian noise injection (max SNR=0.005) to simulate recording artifacts and improve noise robustness, and signal inversion (p=0.2) to enforce polarity invariance—a critical property given that seizure patterns can manifest with opposite polarities across recording sites. 

We define our augmentation function as $
\mathcal{A}: \mathbb{R}^T \rightarrow \mathbb{R}^T
$
where $\mathcal{A}$ is an augmentation operator that transforms a single-channel EEG signal $x(t) \in \mathbb{R}^T$ to an augmented signal $\tilde{x}(t) \in \mathbb{R}^T$, with $T$ being the number of time samples.



The specific augmentation types used are:

 1. Amplitude Scaling
$$
\mathcal{A}_{scale}(x(t)) = \alpha \cdot x(t), \quad \alpha \sim \mathcal{U}(a, b)
$$

 2. Additive Gaussian Noise
$$
\mathcal{A}_{noise}(x(t)) = x(t) + \epsilon(t), \quad \epsilon(t) \sim \mathcal{N}(0, \sigma^2)
$$

 3. Inverting
$$
\mathcal{A}_{invert}(x(t)) = - x(t)
$$



Furthermore, we apply all these augmentations in a probabilistic way such that:
$$
\mathcal{A}_{prob}(x(t)) = \begin{cases}
\mathcal{A}_i(x(t)) & \text{with probability } p_i \\
x(t) & \text{with probability } 1 - \sum_{i} p_i
\end{cases}
$$

Our complete augmentation pipeline is therefore, as follows:

\begin{align}
x_0(t) &= x(t) \quad \text{(original signal)} \\
x_1(t) &= \mathcal{A}_{scale}(x_0(t)) \\
x_2(t) &= \mathcal{A}_{noise}(x_1(t)) \\
x_3(t) &= \mathcal{A}_{invert}(x_2(t)) \\
\end{align}

where 
\vspace{1em}

- $x(t) \in \mathbb{R}^T$: Original single-channel EEG signal

- $\tilde{x}(t) \in \mathbb{R}^T$: Augmented EEG signal  

- $T$: Number of time samples

- $\alpha$: Scaling factor

- $\sigma^2$: Noise variance

- $\mathcal{U}(a,b)$: Uniform distribution between $a$ and $b$

- $\mathcal{N}(\mu, \sigma^2)$: Normal distribution with mean $\mu$ and variance $\sigma^2$
\vspace{1em}

We further divided a full day of data (signal and label) into 20 second epochs (2000 time points each), before using them for training our model.

Ultimately, the combination of residual learning, U-Net's hierarchical temporal representations, and augmentation-based regularization proved essential for achieving robust cross-subject generalization across our 10 test mice at 100Hz temporal resolution.




\subsubsection*{Loss Function}
Dice Loss is now a standard loss function for segmentation tasks across both computer vision and time series applications, making it a natural choice for our seizure detection framework. Originally popularized in medical image segmentation to effectively address class imbalance and spatial overlap measurement\cite{milletari2016v}), Dice Loss has since been successfully adapted to temporal segmentation tasks including sleep stage classification\cite{perslev2019u} and other physiological signal analysis applications. Building on these established successes, we adopted Dice Loss for our cross-subject seizure detection task.


Unlike traditional cross-entropy loss, which treats each pixel or time point independently, Dice Loss considers the global overlap of the segmented regions, making it inherently more robust to class imbalance. This property explains its widespread adoption in both spatial and temporal segmentation tasks. For our application, adopting this approach would not simply predict the majority (non-seizure) class but would instead optimize for accurate detection of the more infrequent event class (i.e. SWDs). 

\subsection*{Model Evaluation}
Due to the highly imbalanced nature of our data--where seizures comprise, at most, 5\% of our dataset--we rely on F1-scores, precision and recall to evaluate the performance of our model, as accuracy would not serve as a reliable metric. Our task involves classifying every time point in the signal as either seizure or non-seizure, and the chosen metrics are appropriate for this class imbalanced problem.

Here, precision is defined as:
$$
\text{Precision} = P = \frac{TP}{TP + FP} = \frac{\text{Correctly predicted positives}}{\text{All predicted positives}}
$$

recall as
$$
\text{Recall} = R = \frac{TP}{TP + FN} = \frac{\text{Correctly predicted positives}}{\text{All actual positives}}
$$

and F1- score as
$$
F_1 = \frac{2 \cdot P \cdot R}{P + R} = \frac{2 \cdot TP}{2 \cdot TP + FP + FN}
$$

where
\[
\begin{aligned}
\mathit{TP} &: \text{True Positives (correctly predicted positive cases)}\\
\mathit{TN} &: \text{True Negatives (correctly predicted negative cases)}\\
\mathit{FP} &: \text{False Positives (incorrectly predicted as positive)}\\
\mathit{FN} &: \text{False Negatives (incorrectly predicted as negative)}
\end{aligned}
\]

\subsubsection*{Comparison models}
\label{subsub:comparison}
To rigorously evaluate the performance of our proposed AugUNet1D architecture, we selected 16 diverse baseline models spanning multiple methodological paradigms and complexity levels. Our selection strategy was designed to provide comprehensive coverage of approaches used in EEG analysis and time series classification tasks. First, we included four traditional machine learning methods, namely Logistic Regression, K-Nearest Neighbors, Decision Tree, and Random Forest, which serve as foundational baselines and represent classical approaches to seizure detection that rely on learned feature representations from the input signals. Second, we incorporated several deep learning architectures of varying complexity: (1) standard CNN and PyramidalCNN for spatial feature extraction; (2) LSTM and bidirectional LSTM (biLSTM) for temporal modeling; and (3) ConvLSTM which combines both spatial and temporal processing. Third, we selected five specialized architectures that have demonstrated success in related physiological signal processing tasks: (1) EEGNet (specifically designed for EEG classification); (2) InceptionTime (a state-of-the-art time series classifier); (3) Xception (an efficient convolutional architecture); (4) vanilla U-Net (to isolate the contribution of our residual and augmentation enhancements); and (5) SalientSleepNet (a recent attention-based architecture for sleep stage classification). Finally, we included DETRtime, a transformer-based object detection model adapted for time series, to assess whether modern query-based detection paradigms are effective for dense temporal segmentation tasks. This diverse baseline selection enables the systematic evaluation of the relative contributions of different architectural components, including encoder-decoder structures, skip connections, residual learning, recurrent processing, and attention mechanisms to performance, and to determine which design principles are most critical for achieving robust cross-subject seizure detection performance.

\subsubsection*{Statistical Comparison of Manual, AugUNet1D, and Twin Peaks Labeling Methods}
Performance scores (F1, recall, precision) of AugUNet1D and Twin Peaks were compared using paired samples t-tests. Differences in features (i.e., duration and peak frequency) among manually labeled SWDs, AugUNet1D-, and Twin Peaks-labeled events were tested using one-way repeated measures ANOVA with detection method as the main factor. Bonferroni post hoc tests were used to compare groups means among all possible combinations of the three detection methods. All statistical tests were performed in GraphPad Prism 8.0.1 (San Diego, CA). Data are reported as mean ± standard deviation (SD) unless otherwise stated. 

\subsubsection*{Training details}

All EEG signals were identically preprocessed. First, raw signals were min-max scaled over their amplitude range to normalize signal magnitudes across subjects. Next, signals were resampled to 100Hz using torchaudio.transforms.Resample to standardize temporal resolution. Finally, the continuous recordings were segmented into 20-second epochs (2000 time points each) to create manageable training examples. The training data from the eight training mice was randomly partitioned into 95\% for model training and 5\% for validation, with the validation set used to monitor performance and select optimal model checkpoints during training. Batch size was 32.

The AugUNet1D model was trained using the Adam optimizer with an initial learning rate of $10^{-3}$ for a maximum of 50 epochs. We employed a CosineAnnealingLinearWarmup learning rate scheduler to facilitate stable training and optimal convergence, which began with a linear warmup phase over the first 500 steps to gradually increase the learning rate from zero to the initial value, preventing early training instability. Following warmup, the scheduler applied cosine annealing over 1000-step cycles with a minimum learning rate of $10^{-5}$ and a decay factor (gamma) of 0.9, allowing the learning rate to oscillate while gradually decreasing over successive cycles to enable fine-tuning of model parameters. To prevent overfitting and reduce unnecessary computational costs, we implemented early stopping with a patience of 10 epochs that monitored the validation Dice loss and terminated training if no improvement was observed for 10 consecutive epochs. This training configuration balanced exploration during early training phases with exploitation during later stages, whereas the early stopping criterion ensured that models were evaluated at their optimal generalization point rather than after potential overfitting. We report results as mean and standard deviation over three runs with random initialization.

For all baseline deep learning methods, we utilized the standardized training code from the DETRtime repository \url{https://github.com/lu-wo/DETRtime/} with necessary modifications to accommodate our mouse EEG dataset and task specifications. Each baseline model was trained for 10 epochs, and the epoch achieving the best validation loss was selected for evaluation on the held-out test set. To account for training variability and ensure robust performance estimates, we trained each baseline model three times with different random initializations and report the mean and standard deviation of precision, recall, and F1-score across these three independent runs. This training protocol ensured fair comparison across all baseline methods while maintaining computational feasibility given the large number of models evaluated.

\section*{Results}

\subsection*{Performance of Machine Learning Methods}

Here, test the performance of a variety of traditional ML and neural network-based methods. 

\begin{figure}[htbp]
\centering
\includegraphics[width=0.9\linewidth]{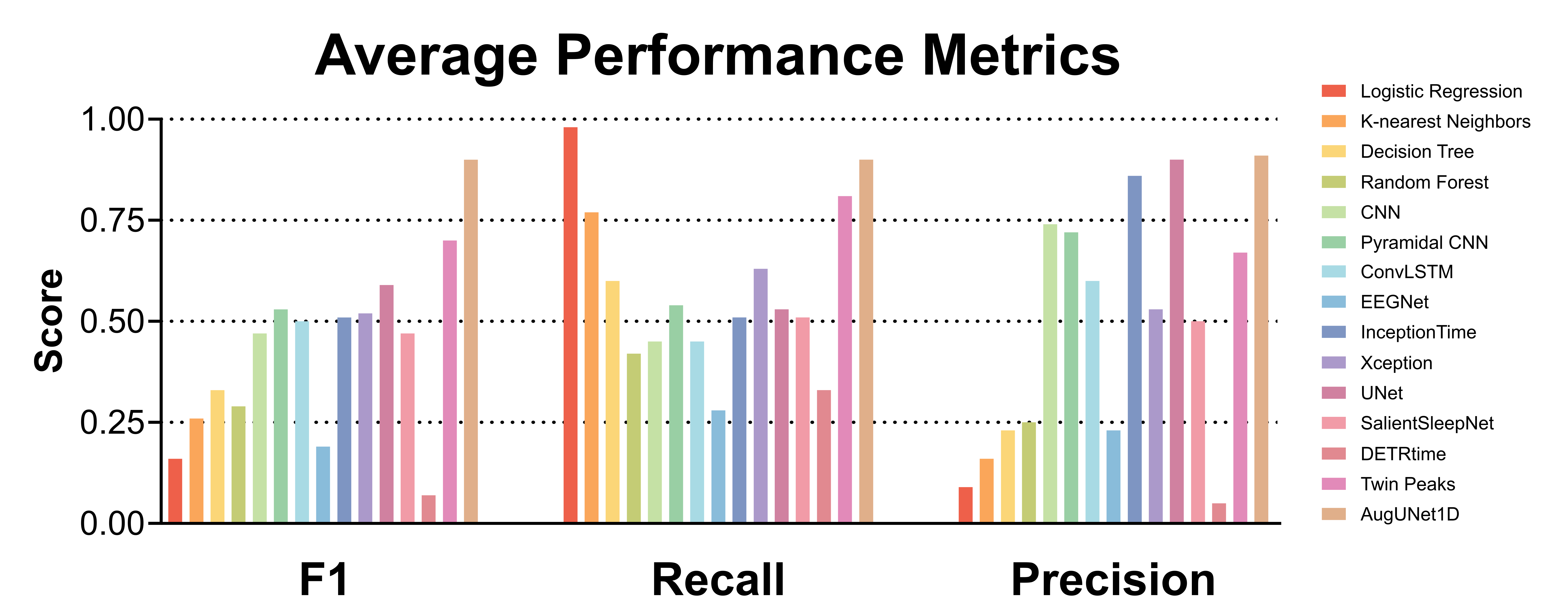}
\caption{Average performance metrics (F1, Recall, Precision) across all recordings in the test dataset. Left-to-right order of bars matches the top-to-bottom order in the legend. Performance by all methods is shown here, except for LSTM and biLSTM, because they yielded zero for all three metrics.}
\label{fig:Fig_PerformanceMetrics}
\end{figure}

We specifically evaluated our proposed AugUNet1D model against 15 baseline methods and the Twin Peaks approach across 10 held-out test mice, using precision, recall, and F1-score as our primary evaluation metrics. The baseline methods spanned four methodological categories: (1) traditional machine learning algorithms (Logistic Regression, K-Nearest Neighbors, Decision Tree, Random Forest); (2) pure deep learning architectures (CNN, PyramidalCNN, LSTM, biLSTM, ConvLSTM); (3) specialized physiological signal processing networks (EEGNet, InceptionTime, Xception, U-Net, SalientSleepNet); and (4) transformer-based detection models (DETRtime). Additionally, we compared AugUNet1D against Twin Peaks, the previous state-of-the-art non-machine learning method on this dataset. Figure \ref{fig:Fig_PerformanceMetrics} shows the models' overall average performance metrics (i.e. F1, recall, and precision scores). Table \ref{tab:results} presents the complete performance comparison across all methods and all 10 test mice. 

\begin{table}[htbp]
\centering
\caption{Table shows the performance metrics for each model tested across all 10 of our test mice. The `avg' columns contains the mean and standard deviation over all mice. We can see that our method performs the best on our dataset.}
\label{tab:results}
\resizebox{\textwidth}{!}{%
\begin{tabular}{lllllllllllll}
\toprule
\textbf{Classifier}                  & \textbf{Metric} & \textbf{M1} & \textbf{M2} & \textbf{M3} & \textbf{M4} & \textbf{M5} & \textbf{M6} & \textbf{M7} & \textbf{M8} & \textbf{M9} & \textbf{M10} & \textbf{avg} \\ 
\midrule
\multirow{3}{*}{Logistic Regression} & precision       & 0.09 ± 0.00 & 0.09 ± 0.00 & 0.10 ± 0.00 & 0.09 ± 0.00 & 0.11 ± 0.00 & 0.10 ± 0.00 & 0.12 ± 0.00 & 0.07 ± 0.00 & 0.07 ± 0.00 & 0.03 ± 0.00  & 0.09 ± 0.00  \\
                                     & recall          & 0.98 ± 0.00 & 0.96 ± 0.00 & 0.97 ± 0.00 & 0.98 ± 0.00 & 0.97 ± 0.00 & 0.98 ± 0.00 & 0.97 ± 0.00 & 0.98 ± 0.00 & 0.98 ± 0.00 & 0.99 ± 0.00  & 0.98 ± 0.00  \\
                                     & f1-score        & 0.16 ± 0.00 & 0.17 ± 0.00 & 0.18 ± 0.00 & 0.16 ± 0.00 & 0.20 ± 0.00 & 0.19 ± 0.00 & 0.22 ± 0.00 & 0.13 ± 0.00 & 0.12 ± 0.00 & 0.06 ± 0.00  & 0.16 ± 0.00  \\ \cline{2-13} 
\multirow{3}{*}{K-Nearest Neighbors} & precision       & 0.15 ± 0.00 & 0.19 ± 0.00 & 0.14 ± 0.00 & 0.15 ± 0.00 & 0.26 ± 0.00 & 0.17 ± 0.00 & 0.25 ± 0.00 & 0.13 ± 0.00 & 0.10 ± 0.00 & 0.06 ± 0.00  & 0.16 ± 0.00  \\
                                     & recall          & 0.70 ± 0.00 & 0.83 ± 0.00 & 0.66 ± 0.00 & 0.79 ± 0.00 & 0.82 ± 0.00 & 0.75 ± 0.00 & 0.71 ± 0.00 & 0.89 ± 0.00 & 0.76 ± 0.00 & 0.83 ± 0.00  & 0.77 ± 0.00  \\
                                     & f1-score        & 0.25 ± 0.00 & 0.31 ± 0.00 & 0.24 ± 0.00 & 0.25 ± 0.00 & 0.39 ± 0.00 & 0.27 ± 0.00 & 0.37 ± 0.00 & 0.23 ± 0.00 & 0.18 ± 0.00 & 0.11 ± 0.00  & 0.26 ± 0.00  \\ \cline{2-13} 
\multirow{3}{*}{Decision Tree}       & precision       & 0.27 ± 0.00 & 0.22 ± 0.00 & 0.25 ± 0.01 & 0.26 ± 0.00 & 0.34 ± 0.01 & 0.25 ± 0.00 & 0.36 ± 0.00 & 0.18 ± 0.00 & 0.14 ± 0.00 & 0.09 ± 0.00  & 0.23 ± 0.00  \\
                                     & recall          & 0.56 ± 0.00 & 0.66 ± 0.00 & 0.45 ± 0.00 & 0.60 ± 0.00 & 0.55 ± 0.01 & 0.60 ± 0.00 & 0.64 ± 0.00 & 0.75 ± 0.01 & 0.62 ± 0.01 & 0.59 ± 0.00  & 0.60 ± 0.00  \\
                                     & f1-score        & 0.36 ± 0.00 & 0.33 ± 0.01 & 0.32 ± 0.01 & 0.36 ± 0.00 & 0.42 ± 0.00 & 0.36 ± 0.00 & 0.46 ± 0.00 & 0.29 ± 0.00 & 0.23 ± 0.00 & 0.16 ± 0.00  & 0.33 ± 0.00  \\ \cline{2-13} 
\multirow{3}{*}{Random Forest}       & precision       & 0.28 ± 0.01 & 0.22 ± 0.01 & 0.24 ± 0.01 & 0.29 ± 0.01 & 0.47 ± 0.01 & 0.20 ± 0.01 & 0.36 ± 0.00 & 0.20 ± 0.00 & 0.13 ± 0.01 & 0.09 ± 0.01  & 0.25 ± 0.01  \\
                                     & recall          & 0.37 ± 0.01 & 0.55 ± 0.02 & 0.26 ± 0.01 & 0.44 ± 0.01 & 0.36 ± 0.02 & 0.32 ± 0.01 & 0.34 ± 0.00 & 0.71 ± 0.02 & 0.46 ± 0.05 & 0.44 ± 0.03  & 0.42 ± 0.02  \\
                                     & f1-score        & 0.32 ± 0.01 & 0.32 ± 0.01 & 0.25 ± 0.01 & 0.35 ± 0.01 & 0.41 ± 0.02 & 0.25 ± 0.01 & 0.35 ± 0.00 & 0.31 ± 0.01 & 0.20 ± 0.02 & 0.15 ± 0.01  & 0.29 ± 0.01  \\  \hline
\multirow{3}{*}{CNN}                 & precision       & 0.66 ± 0.57 & 0.87 ± 0.06 & 0.67 ± 0.58 & 0.67 ± 0.58 & 0.92 ± 0.04 & 0.88 ± 0.03 & 0.84 ± 0.03 & 0.77 ± 0.03 & 0.83 ± 0.02 & 0.33 ± 0.58  & 0.74 ± 0.21  \\
                                     & recall          & 0.01 ± 0.01 & 0.34 ± 0.24 & 0.01 ± 0.01 & 0.01 ± 0.01 & 0.78 ± 0.15 & 0.83 ± 0.04 & 0.86 ± 0.07 & 0.89 ± 0.03 & 0.73 ± 0.06 & 0.00 ± 0.01  & 0.45 ± 0.05  \\
                                     & f1-score        & 0.01 ± 0.02 & 0.46 ± 0.26 & 0.01 ± 0.02 & 0.02 ± 0.02 & 0.84 ± 0.08 & 0.85 ± 0.01 & 0.85 ± 0.02 & 0.83 ± 0.01 & 0.77 ± 0.03 & 0.01 ± 0.01  & 0.47 ± 0.03  \\ \cline{2-13}
\multirow{3}{*}{PyramidalCNN}        & precision       & 0.67 ± 0.58 & 0.88 ± 0.02 & 0.67 ± 0.58 & 0.67 ± 0.58 & 0.86 ± 0.03 & 0.82 ± 0.02 & 0.78 ± 0.01 & 0.73 ± 0.03 & 0.77 ± 0.02 & 0.33 ± 0.58  & 0.72 ± 0.20  \\
                                     & recall          & 0.10 ± 0.09 & 0.63 ± 0.05 & 0.00 ± 0.00 & 0.08 ± 0.08 & 0.92 ± 0.04 & 0.93 ± 0.03 & 0.95 ± 0.02 & 0.95 ± 0.03 & 0.84 ± 0.04 & 0.01 ± 0.01  & 0.54 ± 0.04  \\
                                     & f1-score        & 0.17 ± 0.15 & 0.73 ± 0.03 & 0.01 ± 0.01 & 0.14 ± 0.14 & 0.89 ± 0.00 & 0.87 ± 0.00 & 0.86 ± 0.00 & 0.82 ± 0.01 & 0.81 ± 0.01 & 0.01 ± 0.02  & 0.53 ± 0.03  \\ \cline{2-13}
\multirow{3}{*}{LSTM}                & precision       & 0.00 ± 0.00 & 0.00 ± 0.00 & 0.00 ± 0.00 & 0.00 ± 0.00 & 0.00 ± 0.00 & 0.00 ± 0.00 & 0.00 ± 0.00 & 0.00 ± 0.00 & 0.00 ± 0.00 & 0.00 ± 0.00  & 0.00 ± 0.00  \\
                                     & recall          & 0.00 ± 0.00 & 0.00 ± 0.00 & 0.00 ± 0.00 & 0.00 ± 0.00 & 0.00 ± 0.00 & 0.00 ± 0.00 & 0.00 ± 0.00 & 0.00 ± 0.00 & 0.00 ± 0.00 & 0.00 ± 0.00  & 0.00 ± 0.00  \\
                                     & f1-score        & 0.00 ± 0.00 & 0.00 ± 0.00 & 0.00 ± 0.00 & 0.00 ± 0.00 & 0.00 ± 0.00 & 0.00 ± 0.00 & 0.00 ± 0.00 & 0.00 ± 0.00 & 0.00 ± 0.00 & 0.00 ± 0.00  & 0.00 ± 0.00  \\ \cline{2-13}
\multirow{3}{*}{biLSTM}              & precision       & 0.00 ± 0.00 & 0.00 ± 0.00 & 0.00 ± 0.00 & 0.00 ± 0.00 & 0.00 ± 0.00 & 0.00 ± 0.00 & 0.00 ± 0.00 & 0.00 ± 0.00 & 0.00 ± 0.00 & 0.00 ± 0.00  & 0.00 ± 0.00  \\
                                     & recall          & 0.00 ± 0.00 & 0.00 ± 0.00 & 0.00 ± 0.00 & 0.00 ± 0.00 & 0.00 ± 0.00 & 0.00 ± 0.00 & 0.00 ± 0.00 & 0.00 ± 0.00 & 0.00 ± 0.00 & 0.00 ± 0.00  & 0.00 ± 0.00  \\
                                     & f1-score        & 0.00 ± 0.00 & 0.00 ± 0.00 & 0.00 ± 0.00 & 0.00 ± 0.00 & 0.00 ± 0.00 & 0.00 ± 0.00 & 0.00 ± 0.00 & 0.00 ± 0.00 & 0.00 ± 0.00 & 0.00 ± 0.00  & 0.00 ± 0.00  \\ \cline{2-13}
\multirow{3}{*}{ConvLSTM}            & precision       & 0.00 ± 0.00 & 0.95 ± 0.02 & 0.00 ± 0.00 & 0.33 ± 0.58 & 0.97 ± 0.01 & 0.97 ± 0.02 & 0.95 ± 0.01 & 0.82 ± 0.03 & 0.96 ± 0.02 & 0.00 ± 0.00  & 0.60 ± 0.06  \\
                                     & recall          & 0.00 ± 0.00 & 0.57 ± 0.07 & 0.00 ± 0.00 & 0.00 ± 0.00 & 0.87 ± 0.02 & 0.71 ± 0.07 & 0.81 ± 0.08 & 0.91 ± 0.03 & 0.65 ± 0.07 & 0.00 ± 0.00  & 0.45 ± 0.02  \\
                                     & f1-score        & 0.00 ± 0.00 & 0.71 ± 0.06 & 0.00 ± 0.00 & 0.00 ± 0.00 & 0.92 ± 0.02 & 0.82 ± 0.05 & 0.87 ± 0.05 & 0.86 ± 0.01 & 0.77 ± 0.05 & 0.00 ± 0.00  & 0.50 ± 0.02  \\ \cline{2-13}
\multirow{3}{*}{EEGNet}              & precision       & 0.01 ± 0.02 & 0.54 ± 0.45 & 0.02 ± 0.03 & 0.01 ± 0.02 & 0.54 ± 0.48 & 0.12 ± 0.16 & 0.25 ± 0.38 & 0.62 ± 0.14 & 0.20 ± 0.31 & 0.00 ± 0.01  & 0.23 ± 0.14  \\
                                     & recall          & 0.01 ± 0.02 & 0.18 ± 0.18 & 0.04 ± 0.06 & 0.03 ± 0.06 & 0.58 ± 0.50 & 0.34 ± 0.56 & 0.40 ± 0.53 & 0.85 ± 0.12 & 0.38 ± 0.54 & 0.02 ± 0.03  & 0.28 ± 0.21  \\
                                     & f1-score        & 0.01 ± 0.02 & 0.27 ± 0.25 & 0.03 ± 0.04 & 0.02 ± 0.03 & 0.56 ± 0.49 & 0.05 ± 0.05 & 0.15 ± 0.16 & 0.70 ± 0.06 & 0.09 ± 0.11 & 0.01 ± 0.01  & 0.19 ± 0.10  \\ \cline{2-13}
\multirow{3}{*}{InceptionTime}       & precision       & 0.85 ± 0.25 & 0.88 ± 0.06 & 0.67 ± 0.58 & 1.00 ± 0.00 & 0.89 ± 0.04 & 0.88 ± 0.02 & 0.84 ± 0.02 & 0.72 ± 0.02 & 0.82 ± 0.02 & 1.00 ± 0.00  & 0.86 ± 0.05  \\
                                     & recall          & 0.03 ± 0.04 & 0.62 ± 0.13 & 0.00 ± 0.01 & 0.04 ± 0.06 & 0.90 ± 0.04 & 0.88 ± 0.03 & 0.92 ± 0.02 & 0.94 ± 0.02 & 0.79 ± 0.02 & 0.01 ± 0.00  & 0.51 ± 0.03  \\
                                     & f1-score        & 0.05 ± 0.07 & 0.72 ± 0.08 & 0.01 ± 0.01 & 0.07 ± 0.11 & 0.89 ± 0.01 & 0.88 ± 0.02 & 0.88 ± 0.02 & 0.81 ± 0.01 & 0.81 ± 0.02 & 0.01 ± 0.00  & 0.51 ± 0.03  \\ \cline{2-13}
\multirow{3}{*}{Xception}            & precision       & 0.02 ± 0.03 & 0.94 ± 0.01 & 0.02 ± 0.03 & 0.02 ± 0.03 & 0.92 ± 0.02 & 0.90 ± 0.04 & 0.87 ± 0.04 & 0.75 ± 0.05 & 0.87 ± 0.06 & 0.00 ± 0.01  & 0.53 ± 0.02  \\
                                     & recall          & 0.31 ± 0.54 & 0.59 ± 0.07 & 0.31 ± 0.54 & 0.31 ± 0.54 & 0.91 ± 0.01 & 0.89 ± 0.05 & 0.94 ± 0.02 & 0.96 ± 0.01 & 0.82 ± 0.03 & 0.30 ± 0.52  & 0.63 ± 0.22  \\
                                     & f1-score        & 0.03 ± 0.05 & 0.72 ± 0.05 & 0.04 ± 0.06 & 0.03 ± 0.06 & 0.92 ± 0.01 & 0.90 ± 0.01 & 0.90 ± 0.02 & 0.84 ± 0.03 & 0.84 ± 0.01 & 0.01 ± 0.02  & 0.52 ± 0.02  \\ \hline
\multirow{3}{*}{U-Net}                & precision       & 0.99 ± 0.01 & 0.62 ± 0.51 & 1.00 ± 0.00 & 0.99 ± 0.01 & 0.95 ± 0.02 & 0.96 ± 0.01 & 0.94 ± 0.01 & 0.57 ± 0.11 & 0.94 ± 0.00 & 0.98 ± 0.02  & 0.90 ± 0.05  \\ 
                                     & recall          & 0.26 ± 0.19 & 0.34 ± 0.08 & 0.20 ± 0.27 & 0.40 ± 0.26 & 0.86 ± 0.06 & 0.69 ± 0.13 & 0.75 ± 0.12 & 0.97 ± 0.03 & 0.61 ± 0.11 & 0.24 ± 0.31  & 0.53 ± 0.13  \\
                                     & f1-score        & 0.38 ± 0.24 & 0.39 ± 0.28 & 0.29 ± 0.34 & 0.54 ± 0.26 & 0.90 ± 0.02 & 0.80 ± 0.09 & 0.83 ± 0.08 & 0.71 ± 0.09 & 0.74 ± 0.08 & 0.32 ± 0.38  & 0.59 ± 0.15  \\ \cline{2-13}
\multirow{3}{*}{SalientSleepNet}     & precision       & 0.27 ± 0.46 & 0.60 ± 0.41 & 0.27 ± 0.47 & 0.25 ± 0.43 & 0.86 ± 0.09 & 0.55 ± 0.48 & 0.85 ± 0.13 & 0.66 ± 0.01 & 0.52 ± 0.45 & 0.21 ± 0.37  & 0.50 ± 0.25  \\
                                     & recall          & 0.28 ± 0.48 & 0.44 ± 0.37 & 0.30 ± 0.52 & 0.31 ± 0.53 & 0.69 ± 0.45 & 0.61 ± 0.53 & 0.64 ± 0.55 & 0.95 ± 0.00 & 0.55 ± 0.48 & 0.30 ± 0.51  & 0.51 ± 0.39  \\
                                     & f1-score        & 0.27 ± 0.47 & 0.50 ± 0.41 & 0.28 ± 0.49 & 0.27 ± 0.47 & 0.67 ± 0.34 & 0.58 ± 0.50 & 0.57 ± 0.49 & 0.78 ± 0.01 & 0.53 ± 0.46 & 0.25 ± 0.43  & 0.47 ± 0.36  \\ \cline{2-13}
\multirow{3}{*}{DETRtime}            & precision       & 0.06 ± 0.01 & 0.05 ± 0.01 & 0.04 ± 0.04 & 0.04 ± 0.02 & 0.06 ± 0.01 & 0.07 ± 0.01 & 0.06 ± 0.03 & 0.02 ± 0.02 & 0.04 ± 0.01 & 0.01 ± 0.00  & 0.05 ± 0.01  \\
                                     & recall          & 0.34 ± 0.28 & 0.32 ± 0.27 & 0.36 ± 0.31 & 0.33 ± 0.28 & 0.31 ± 0.25 & 0.36 ± 0.30 & 0.35 ± 0.29 & 0.31 ± 0.27 & 0.33 ± 0.26 & 0.33 ± 0.28  & 0.33 ± 0.28  \\
                                     & f1-score        & 0.07 ± 0.04 & 0.07 ± 0.04 & 0.08 ± 0.07 & 0.07 ± 0.05 & 0.08 ± 0.04 & 0.09 ± 0.05 & 0.10 ± 0.07 & 0.04 ± 0.04 & 0.05 ± 0.01 & 0.02 ± 0.01  & 0.07 ± 0.04  \\ \hline
\multirow{3}{*}{Twin Peaks}    & precision & 0.57 ± 0.00 & 0.96 ± 0.00 & 0.59 ± 0.00 & 0.64 ± 0.00 & 0.82 ± 0.00 & 0.82 ± 0.00 & 0.89 ± 0.00 & 0.39 ± 0.00 & 0.71 ± 0.00 & 0.25 ± 0.00 & 0.67 ± 0.00            \\
                               & recall    & 0.73 ± 0.00 & 0.45 ± 0.00 & 0.83 ± 0.00 & 0.92 ± 0.00 & 0.82 ± 0.00 & 0.89 ± 0.00 & 0.86 ± 0.00 & 0.92 ± 0.00 & 0.81 ± 0.00 & 0.95 ± 0.00 & 0.82 ± 0.00            \\
                               & f1-score  & 0.64 ± 0.00 & 0.62 ± 0.00 & 0.69  ± 0.00 & 0.75 ± 0.00 & 0.81  ± 0.00 & 0.85 ± 0.00 & 0.87 ± 0.00 & 0.55 ± 0.00 & 0.76 ± 0.00 & 0.40 ± 0.00 & 0.69 ± 0.00     \\ \hline
\multirow{3}{*}{AugUNet1D (ours)}    & precision       & 0.92 ± 0.01 & 0.96 ± 0.01 & 0.92 ± 0.01 & 0.87 ± 0.01 & 0.97 ± 0.02 & 0.95 ± 0.01 & 0.94 ± 0.02 & 0.83 ± 0.02 & 0.91 ± 0.03 & 0.79 ± 0.03  & 0.91 ± 0.02  \\
                                     & recall          & 0.84 ± 0.01 & 0.66 ± 0.09 & 0.94 ± 0.00 & 0.97 ± 0.00 & 0.89 ± 0.02 & 0.95 ± 0.00 & 0.96 ± 0.01 & 0.95 ± 0.00 & 0.89 ± 0.02 & 0.96 ± 0.01  & 0.90 ± 0.02  \\
                                     & f1-score        & 0.87 ± 0.00 & 0.78 ± 0.06 & 0.93 ± 0.01 & 0.92 ± 0.01 & 0.93 ± 0.00 & 0.95 ± 0.00 & 0.95 ± 0.01 & 0.89 ± 0.01 & 0.90 ± 0.01 & 0.87 ± 0.02  & 0.90 ± 0.01  \\
\bottomrule
\end{tabular}%
}
\end{table}

Table 2 shows that our proposed AugUNet1D performed best, with an average F1-score of 0.90 ± 0.01, representing a 29\% relative improvement over the Twin Peaks baseline (0.69 F1-score) and a 70\% relative improvement over the best general deep learning baseline. Notably, AugUNet1D demonstrated exceptional cross-subject consistency with standard deviations of only 0.01-0.02 across all metrics, the lowest variance among all evaluated methods. The model achieved robust individual performance across all test subjects, with F1-scores ranging from 0.87 to 0.93, indicating successful generalization to held out mice. Furthermore, AugUNet1D maintained a balanced precision-recall trade-off with 0.91 precision and 0.90 recall, compared to Twin Peaks' 0.67 precision and 0.81 recall, suggesting that our model achieves superior detection accuracy without sacrificing either sensitivity or specificity.

The baseline methods exhibited dramatically different performance profiles that reveal important insights about architectural requirements for cross-subject seizure detection. Traditional machine learning methods performed poorly across the board, with F1-scores ranging from 0.16 for Logistic Regression to 0.33 for Decision Tree. Despite their consistent performance (low variance), these methods fundamentally struggled with the cross-subject temporal segmentation task likely due to their limited capacity to learn complex hierarchical representations from raw EEG signals. The failure of these methods underscores the necessity of deep learning approaches to solve this challenging problem.

Surprisingly, pure recurrent architectures including LSTM, biLSTM, and ConvLSTM completely failed to learn effective representations, achieving 0.00 F1-scores across all test mice. This catastrophic failure suggests that recurrent networks, despite their theoretical suitability for sequential data, require careful initialization and architectural support to handle cross-subject generalization during high-frequency EEG signals. By contrast, standard convolutional architectures achieved moderate success, with vanilla CNN reaching a 0.47 ± 0.05 F1-score and PyramidalCNN slightly improving to an F1-score of 0.53 ± 0.04. However, these models exhibited high variance in precision (0.74 ± 0.21 for CNN), suggesting inconsistent performance across different subjects. The modest improvement of PyramidalCNN over vanilla CNN indicates that hierarchical feature extraction provides some benefit, but still falls far short for robust cross-subject detection.

Among specialized architectures, vanilla U-Net achieved the highest F1-score of 0.59 ± 0.15, emerging as the strongest general deep learning baseline. However, its high standard deviation of 0.15 reveals significant instability across subjects, highlighting the importance of our residual connection modifications. InceptionTime and Xception both achieved identical performance, demonstrating that multi-scale temporal feature extraction and efficient convolutional designs alone are insufficient to overcome the cross-subject generalization challenge. Most surprisingly, EEGNet achieved only 0.19 ± 0.10 F1-score despite being specifically designed for EEG classification tasks. SalientSleepNet, despite incorporating attention mechanisms, achieved 0.47 ± 0.36 F1-score with extremely high variance (0.36 standard deviation), indicating that attention alone does not guarantee robust cross-subject generalization. 


The transformer-based DETRtime model failed dramatically with only 0.07 ± 0.04 F1-score. We hypothesize that query-based object detection paradigms designed for sparse event detection are fundamentally mismatched with dense point-wise temporal segmentation tasks. This result validates our architectural choice of encoder-decoder networks with skip connections over modern detection-based approaches for continuous seizure monitoring applications.


\subsection*{Performance Comparison between AugUNet1D and Twin Peaks}
Recently, "Twin Peaks" was developed for detecting SWDs generated by the WAG/Rij rat model of absence epilepsy, an approach that showed strong performance\cite{iotchev2024twin}. To determine how AugUNet1D compares to Twin Peaks at a granular resolution, we applied both methods to our test dataset and compared performance metrics as well as features of detected events (i.e. durations and peak frequencies). 

In these comparisons, performance was measured on a whole event basis. That is, if a detected event had any overlap with a manually labeled event, it was considered a true positive. We classified events with no such overlap at all as false positives (Figure 6A). Similarly, only manually labeled events that were missed entirely were considered false negatives (Figure 6B). Using this approach AugNet1D had a significantly higher mean F1 score (0.95 ± 0.04) than Twin Peaks (0.71 ± 0.15; \textit{t}\textsubscript{(9)} = 5.34, \textit{p} < 10\textsuperscript{-3}). AugNet1D also showed higher precision (0.96 ± 0.03) than Twin Peaks (0.58 ± 0.18; \textit{t}\textsubscript{(9)} = 7.38, \textit{p} < 10\textsuperscript{-4}). Twin Peaks exhibited a slightly higher recall (0.96 ± 0.07) than AugUNet1D (0.94 ± 0.06; \textit{t}\textsubscript{(9)} = 3.95, \textit{p} = 0.003), but the magnitude of the effects observed for F1 and precision scores were far larger. Because F1 scores balance precision and recall, we conclude that AugUNet1D outperformed Twin Peaks in our test dataset, having better F1 scores in every recording therein (Figure 6C).

\begin{figure}[htbp]
\centering
\includegraphics[width=0.7\linewidth]{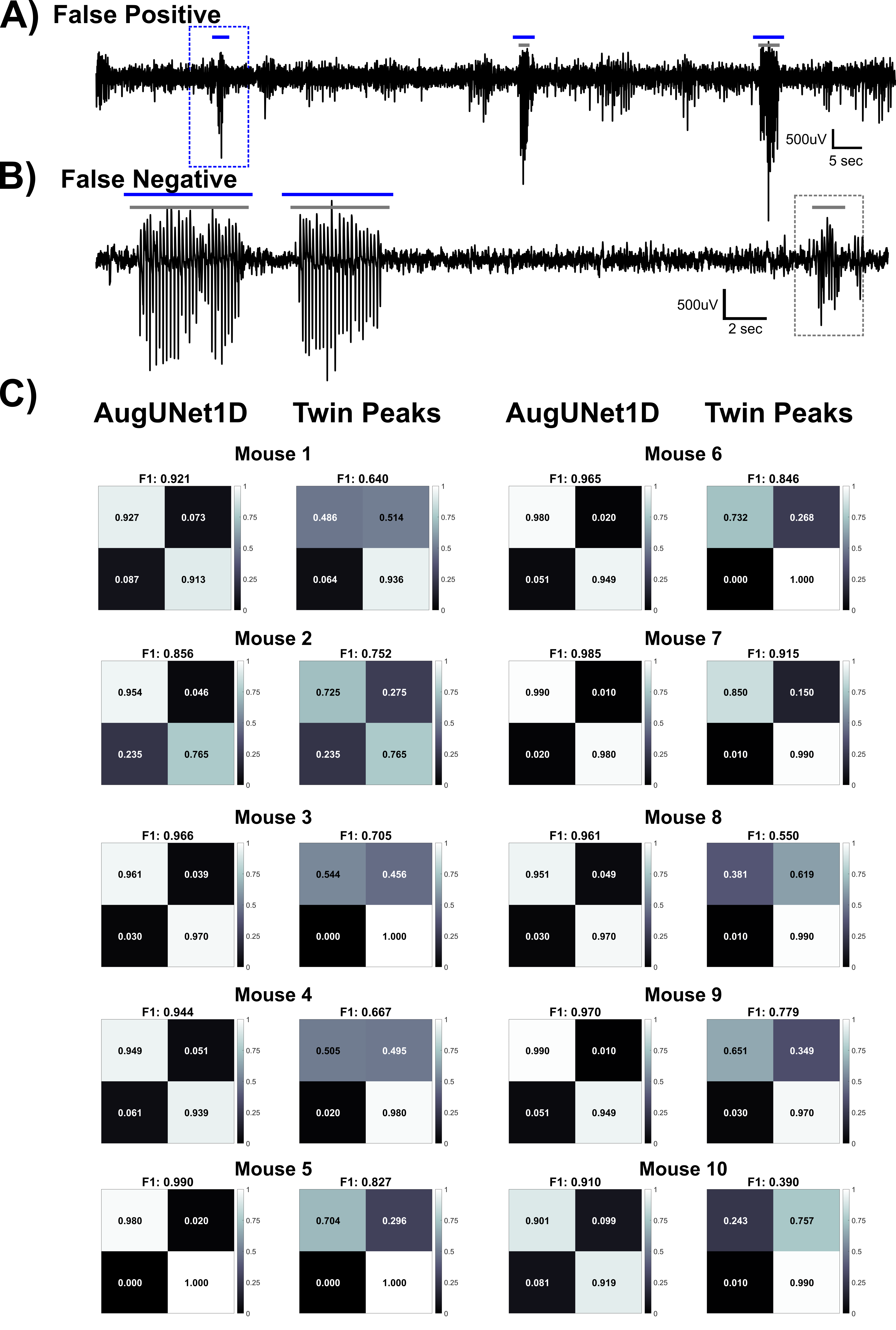}
\caption{False positives, negatives, confusion matrices and F1 scores for AugUNet1D and Twin Peaks. A) Example of false positive in the dotted blue box. B) Example of false negative in the dotted gray box. Blue lines correspond to events detected by AugUNet1D. Gray lines correspond to manually labeled events in A and B. C) Confusion matrices and F1 scores for AugUNet1D and Twin Peaks prediction approaches for each mouse in the test dataset. Proportions in colored panels are as follows: Top left, true positives to all positives. Top right, false positives to all positives. Bottom left, false negatives to all negatives. Bottom right, true negatives to all negatives. These values were calculated on a per event basis, meaning that, for example, a predicted SWD was considered true if it had any overlap with an manually labeled SWD and false otherwise.}
\label{fig:Fig_ConfusionMatrices}
\end{figure}

We also systematically compared features of events detected by AugUNet1D and Twin Peaks to determine if either method has a bias towards particular event types. Specifically, we compared the peak frequency of detected events and the average duration of events, which, in our original training data, were similar across time and across mice. A one-way repeated measures ANOVA revealed a significant effect of detection method on event duration (F\textsubscript{(2,18)} = 92.7, \textit{p} < 10\textsuperscript{-4}; Figure 7A). Post hoc comparison showed that Twin Peaks detected events that were significantly briefer (2.35 ± 0.74s) than those detected either manually (4.88 ± 0.89s; t\textsubscript{(9)} = 11.7, \textit{p} < 10\textsuperscript{-3}) or by AugUNet1D (4.93 ± 0.82s; t\textsubscript{(18)} = 11.9, \textit{p} < 10\textsuperscript{-3}; Figure 7B). One-way repeated measures ANOVA also showed a significant effect of detection method on peak frequencies (F\textsubscript{(2,18)} = 18.25, \textit{p} < 10\textsuperscript{-4}; Figure 7C). Twin Peaks detected events with higher frequencies (6.58 ± 0.42Hz) than those detected by AugUNet1D (6.13 ± 0.14Hz; t\textsubscript{(18)} = 5.26, \textit{p} < 10\textsuperscript{-3}) or manually (6.14 ± 0.13Hz; t\textsubscript{(9)} = 5.2, \textit{p} < 10\textsuperscript{-3}; Figure 7D). There were no significant differences between AugUNet1D and manual detection in event duration (\textit{t}\textsubscript{(18)} = 0.2, \textit{p} > 0.99) or peak frequency (\textit{t}\textsubscript{(18)} = 0.06, \textit{p} > 0.99). To summarize, in addition to performing better in terms of overall F1 score, AugUNet1D also detected events with features similar to manually labeled SWDs. By contrast, Twin Peaks detected events with higher peak frequencies and shorter durations.

\begin{figure}[htbp]
\centering
\includegraphics[width=0.9\linewidth]{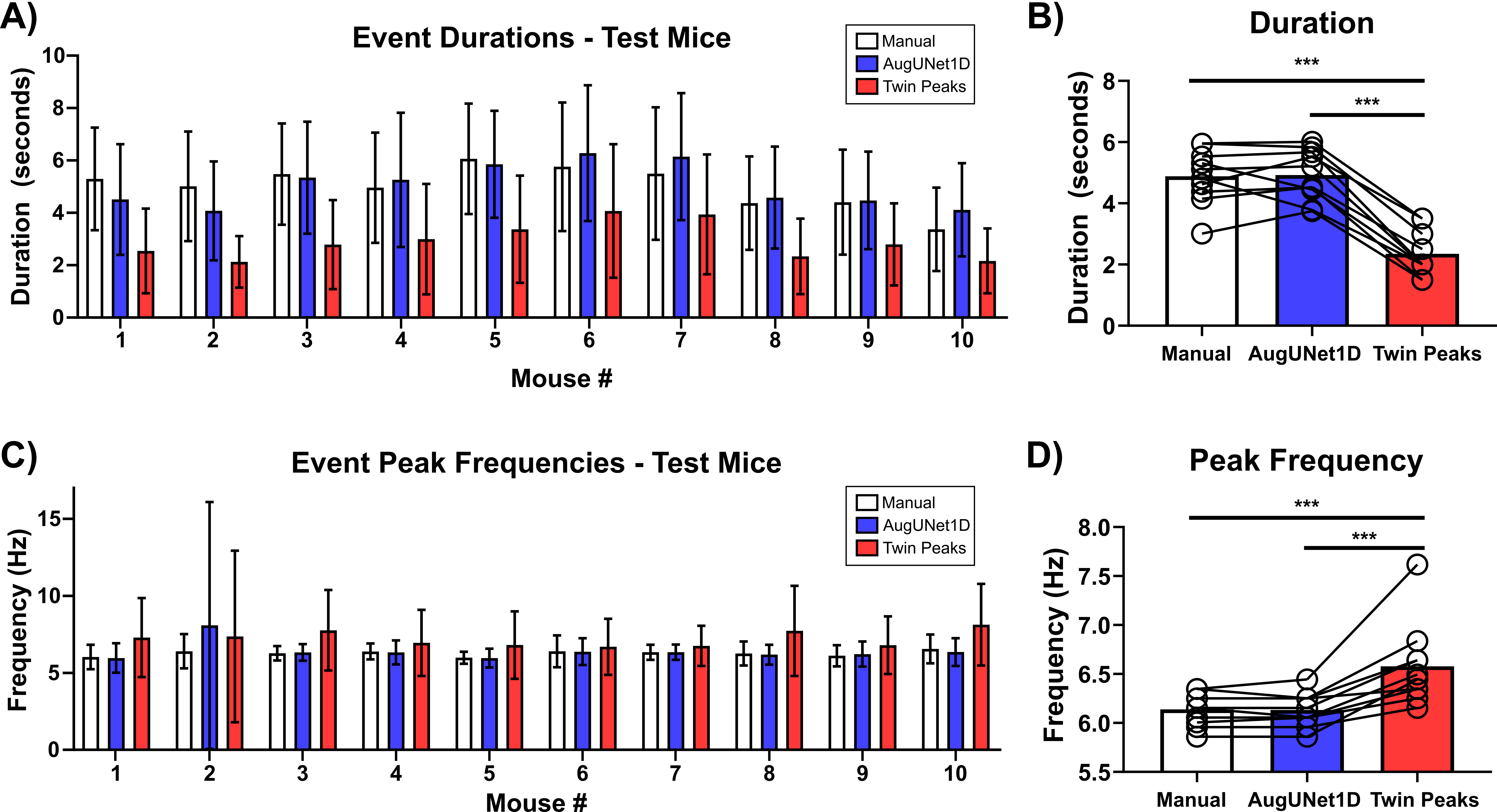}
\caption{Twin Peaks method shows a bias for shorter events with higher peak frequencies. A) Mean durations for events detected manually, with AugUNet1D, or with Twin Peaks in all ten mice in the test dataset. B) Bar heights represent mean durations across all mice. Open circles show the data for each mouse. C and D) Same as A and B but for peak frequencies of detected events. Error bars show ± standard deviation. *** p < 10\textsuperscript{-3}}
\label{fig:Fig_TestDataStats}
\end{figure}

\subsection*{Event Detection during Noise and Sleep}
We next asked whether events were erroneously detected during sleep or during electrographic noise induced by movement, poor mechanical connection, cable swinging, or during sleep. Briefly, noise epochs were defined by blocks of time filled with artificially high amplitude EEG signal (see Materials and Methods, Figure 8A). These recordings had very few noise epochs overall, making it difficult to estimate how these three methods performed during those periods. One-way repeated measures ANOVA revealed no significant effect of detection method on the proportion of events during noise (F\textsubscript{(2,18)} = 2.411, \textit{p} = 0.214; Figure 8B). 

Sleep epochs were identified by an increased amplitude in the 0.1 to 4Hz-bandpass filtered EEG signal (see Materials and Methods; Figure 8C). One-way repeated measures ANOVA revealed a significant effect of detection method on the proportion of events during sleep (F\textsubscript{(2,18)} = 35.06, \textit{p} < 10\textsubscript{-4}). Post hoc comparisons showed that a greater proportion of events during sleep were detected by Twin Peaks (0.46 ± 0.15) than by manual labeling (0.17 ± 0.07; \textit{t}\textsubscript{18} = 7.36 \textit{p} < 10\textsubscript{-4}) and AugUNet1D (0.18 ± 0.017; \textit{t}\textsubscript{18} = 7.14 \textit{p} < 10\textsubscript{-4}). There was no difference between manually detected and AugUNet1D-detected events (\textit{t}\textsubscript{(18)} = 0.22, p > 0.99, Figure 8D). In conclusion, short, non-SWD events are erroneously detected by Twin Peaks during sleep. 

\begin{figure}[htbp]
\centering
\includegraphics[width=0.9\linewidth]{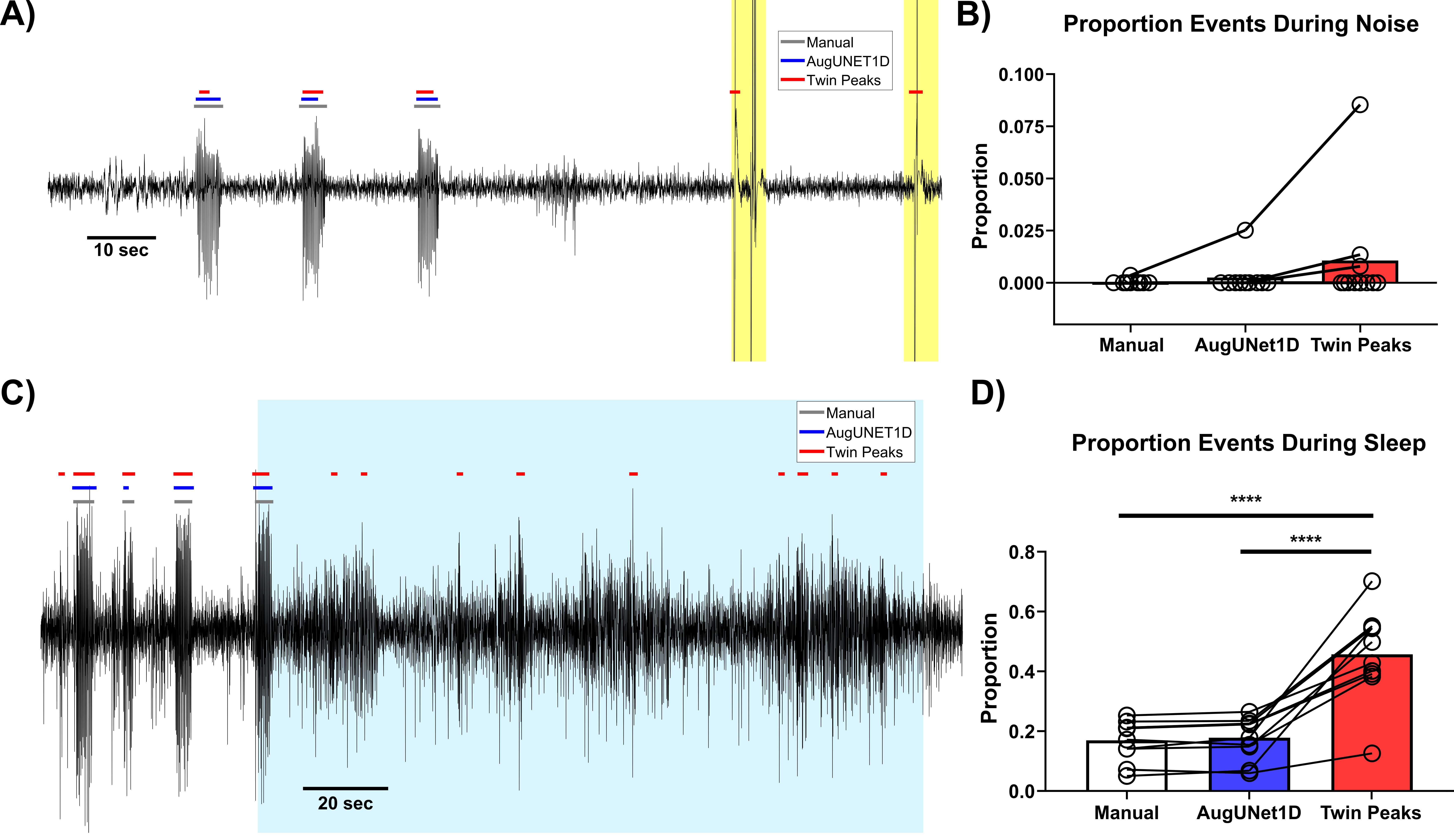}
\caption{Cross-method comparison of detected events during noise and sleep. A) Example of the Twin Peaks method falsely detecting two events during periods with high-amplitude cable artifact highlighted in yellow shading. B) Overall, very few events were detected during noise epochs, but Twin Peaks detected more than AugUNet1D and manual labeling. B) The proportion of events that Twin Peaks detects during sleep is higher than the number of manually detected events and those detected by AugUNet1D. B) Example trace with Twin Peaks falsely detecting many events during sleep which is indicated in blue shading D) Twin Peaks detects many more events during sleep than manual labeling and AugUNet1D. **** p < 10\textsuperscript{-4}}
\label{fig:Fig_StateEvents}
\end{figure}

\section*{Ablation studies}


\subsubsection*{Results of different augmentations}

AugUNet1D uses several data augmentation strategies during training to diversify the dataset and improve generalization to new, unseen data. We conducted an ablation study to assess the individual and combined contributions of each data augmentation strategy: Gaussian noise injection, signal inversion, and amplitude scaling—on the cross-subject seizure detection performance of our residual 1D U-Net architecture. The results reveal critical insights into the role of augmentation in achieving robust generalization. Results are presented in Table \ref{tab:augments}. 

\begin{table}[h]
\centering
\caption{\label{tab:augments}Model performance for different chosen augmentations. We can see scaling provides the most benefit for our use case.}
\begin{tabular}{lrrr}
\toprule
\textbf{Augmentation} & \multicolumn{1}{l}{\textbf{Test Precision}} & \multicolumn{1}{l}{\textbf{Test Recall}} & \multicolumn{1}{l}{\textbf{Test F1}} \\
\midrule
No Augmentation       & 0.9464                               & 0.2755                                      & 0.4268                                      \\
Gaussian Noise             & 0.9544                               & 0.3116                                      & 0.4698                                      \\
Invert        & 0.9591                               & 0.3631                                      & 0.5267                                      \\
Scaling               & 0.8783                               & 0.8442                                      & 0.8609                                      \\
All augmentations     & 0.9092                               & 0.8618                                      & 0.8848                     \\
\bottomrule
\end{tabular}

\end{table}

Without any data augmentation, the model achieved high precision (0.9464) but catastrophically low recall (0.2755), resulting in a poor F1-score of 0.4268. This severe imbalance indicates that the unaugmented model learned to be extremely conservative in its predictions, likely defaulting to predicting the majority (non-seizure) class to minimize false positives. Although the model could identify seizures with high confidence when it made positive predictions, it missed the vast majority of bona fide seizure events, rendering it clinically ineffective.

Including Gaussian noise augmentation alone into the model provided only marginal improvement, increasing recall to 0.3116 and the F1-score to 0.4698 while maintaining high precision (0.9544). This modest benefit suggests that noise injection helps the model become slightly more robust to signal variations, but does not fundamentally address the class imbalance problem. Signal inversion augmentation showed slightly greater impact, improving recall to 0.3631 and F1-score to 0.5267 with precision of 0.9591. 

By contrast, amplitude scaling augmentation alone produced a dramatic transformation in model behavior, achieving a recall of 0.8442 and F1-score of 0.8609. While precision decreased moderately to 0.8783, the substantial gain in recall demonstrates that scaling augmentation is by far the most critical component for cross-subject generalization. 

Ultimately, combing all three augmentation strategies yielded the best overall performance with an F1-score of 0.8848, achieving an optimal balance between precision (0.9092) and recall (0.8618). The combined approach outperformed even amplitude scaling alone, indicating that while scaling addresses the primary challenge of amplitude variability, the additional regularization resulting from noise injection and the polarity invariance from signal inversion provide complementary benefits that further improve model robustness. 


\subsubsection*{Results of increasing percentage of training data}

Because AugUNet1D performs well when labeling SWDs, we were motivated to evaluate model perforamnce when trained with fewer events (our training dataset was quite large: 22,637 SWDs, 173,160 20-second segments). To estimate the quantity of training data needed to achieve adequate performance, we investigated the data efficiency of our proposed AugUNet1D model by training on varying fractions of the available training data (5\%, 10\%, 25\%, 50\%, 75\%, and 90\%), and then evaluating performance on the same held-out test set. Our results (Table \ref{tab:fraction_performance}) demonstrate remarkable data efficiency while revealing important insights about the model's learning dynamics and the role of training data quantity in cross-subject generalization.

\begin{table}[htbp]
\centering
\caption{Table showing the performance of the proposed model with different fractions of the training data. The total number of labeled 20-second segments is 173160.}
\label{tab:fraction_performance}
\begin{tabular}{lllll}
\toprule
\textbf{Training Data Percentage} & \textbf{\# Labeled Segments} & \textbf{Test Precision} & \textbf{Test Recall} & \textbf{Test F1} \\ \midrule
5\%                     & 8658 & 0.9350                   & 0.7289                & 0.8192            \\
10\%                     & 17316 & 0.9353                   & 0.7166                & 0.8115            \\
25\%                     & 43290 & 0.9281                   & 0.7913                & 0.8543            \\
50\%                     & 86580 & 0.9214                   & 0.7956                & 0.8538            \\
75\%                     & 129870 & 0.9184                   & 0.7952                & 0.8524            \\
90\%                     & 155844 & 0.9331                   & 0.8007                & 0.8618            \\
\bottomrule
\end{tabular}%
\end{table}

By limiting the training dataset to only 5\%, the model achieved surprisingly strong performance with an F1-score of 0.8192, maintaining high precision (0.9350) and reasonable recall (0.7289). This result indicates that the combination of our residual U-Net architecture and comprehensive augmentation strategy enables effective learning from limited labeled data, a potentially critical advantage for seizure detection applications where obtaining large, annotated datasets is excessively laborious, expensive, or time-consuming. 
Performance improvements were most pronounced when increasing training data from 10\% to 25\%, with the F1-score increasing to 0.8543. This substantial gain indicates that while the model can learn basic seizure patterns from minimal data, exposure to a more diverse set of training examples significantly enhances generalization. 

Beyond 50\% of training data, performance improvements plateaued, with 75\% and 90\% of data yielding F1-scores of 0.8524 and 0.8618, respectively. Notably, evaluating the precision-recall dynamics across different training data fractions revealed interesting patterns. Precision remained remarkably stable across all conditions (0.9184-0.9353), whereas recall showed  primary improvements as training data increased (0.7166 at 10\% to 0.8007 at 90\%). This pattern indicates that the model maintains consistent specificity regardless of training set size but requires more diverse training examples to improve its sensitivity to varied seizure manifestations.



These findings have important practical implications for the deployment of seizure detection systems. The strong performance achieved with only 25-50\% of training data suggests that effective models can be developed without requiring exhaustive data collection from large numbers of subjects. However, the plateau beyond 50\% also indicates that simply collecting more data from the same subject population may not substantially improve cross-subject performance. Instead, improvements may require either architectural innovations, more sophisticated augmentation strategies, or the inclusion of training data from more diverse subject populations with varied seizure characteristics.





\subsubsection*{Results of different values of probability of applying amplitude scaling}

Finally, we conducted a hyperparameter study to resolve the optimal probability (p) for applying amplitude scaling augmentation during training, evaluating probabilities ranging from 0.1 to 0.5. Our results (\ref{tab:prob_scaling}) revealed a critical trade-off between precision and recall that is directly influenced by the augmentation probability, with substantial implications for model robustness and clinical utility.

The results are available in Table \ref{tab:prob_scaling}.

\begin{table}[htbp]
\centering
\caption{}
\label{tab:prob_scaling}
\begin{tabular}{llll}
\toprule
\textbf{Scale (p)} & \textbf{Test Precision} & \textbf{Test Recall} & \textbf{Test F1} \\ \midrule
0.1                & 0.9451                   & 0.7680                & 0.8474            \\
0.2                & 0.9239                  & 0.8003               & 0.8577           \\
0.3                & 0.9435                  & 0.7602               & 0.8420           \\
0.4                & 0.9378                  & 0.7628               & 0.8413           \\
0.5                & 0.9092                  & 0.8618               & 0.8840            \\ \bottomrule
\end{tabular}%
\end{table}

At lower augmentation probabilities (p=0.1, 0.3, and 0.4), the model exhibited consistently high precision (0.9378-0.9451) but relatively modest recall (0.7602-0.7680), resulting in F1-scores between 0.841 and 0.847. This pattern indicates that when amplitude scaling is applied infrequently during training, the model learns conservative decision boundaries that prioritize specificity over sensitivity. While such models make few false positive predictions, they fail to detect approximately 23-24\% of actual seizure events, limiting their clinical utility for patient monitoring applications where missing seizures can have serious consequences.

A notable exception emerged at p=0.2, which achieved competitive performance with an F1-score of 0.8577 through a more balanced precision-recall profile (0.9239 precision, 0.8003 recall). This observation suggests that a moderate augmentation frequency provides some improvement in recall while maintaining strong precision. However, increasing the augmentation probability to p=0.5 produced the optimal performance with an F1-score of 0.884, representing the best overall detection capability. This configuration achieved a substantial gain in recall to 0.8618 while maintaining strong precision at 0.9092, successfully detecting approximately 86\% of seizure events with 91\% of positive predictions being correct.

\section*{Discussion}

Reliably and accurately detecting events of interest in EEG signals is a common problem. Manually scanning through EEG traces and demarcating relevant intervals is time intensive and can be affected by extraneous factors like scorer experience, bias, and attention level. Attempts to automatically detect salient EEG events often involve using features in the time and frequency domains of the EEG (e.g. spectral profiles, changes in powers of specific frequency bands, waveform shape)\cite{van2008automatic,ovchinnikov2010algorithm,ozmen2021automatic,iotchev2024twin}, machine learning approaches \cite{jando1993pattern,navas2024machine}, or a combination of both \cite{pfammatter2019automated,baser2022automatic}. However, when detecting frequent, spontaneous SWDs observed during long continuous recordings from C3H/HeJ mice, we found available methods inadequate. Inspired by U-Nets used in image segmentation and classification\cite{ronneberger2015u,komura2018machine,yuan2024medical}, we applied the same neural network architecture to the problem of SWD detection. The result is AugUnet1D, a U-Net neural network architecture for 1-dimensional time series data that uses data augmentation strategies during training to optimize SWD detection. 

After training the network with 961 hours of EEG recording, including 22,637 SWDs labeled manually by experienced researchers, AugUNet1D was tested on 10 EEG recordings from 10 different mice not included in the original training dataset. It achieved an overall average F1 score of 0.90 with a recall of 0.89 and precision of 0.90 (\ref{tab:results}). By comparison with all other models, it is clear that the combination of AugUNet1D's encoder-decoder architecture with skip connections, residual learning for gradient stability, and data augmentation significantly improve performance on cross-subject seizure detection. The U-Net architecture implemented in AugUNet1D permits the network to create hierarchical temporal representations, which explain much of its success on this data labeling task. Traditional machine learning approaches like Decision Tree and Logistic Regression performed far worse than AugUNet1D, likely due to their limited capacity to learn complex hierarchical representations from raw EEG signals. Recurrent architectures including LSTM, biLSTM, and ConvLSTM also failed despite their theoretical suitability for sequential data. The inability of these models to produce any meaningful predictions indicates that temporal modeling alone, without spatial feature extraction through convolution, is insufficient for this task. Other architectures (e.g., EEGNet) likely fail because they were designed for multi-channel spatial-spectral feature extraction which differs fundamentally from the dense temporal segmentation required for SWD detection. Finally, a recently published time- and frequency-based method, Twin Peaks, was also inferior to AugUNet1D and has lower temporal resolution. AugUNet then represents the best overall method for SWD detection for our dataset.

In addition to performing better overall, one especially important advantage of AugUNet1D is that it can very precisely segment SWDs. It marks the starts and ends of detected events very close to where trained researchers draw event boundaries (see Figure 8 for examples). Time- and frequency-based detection methods rely on changes in the power spectrum, which has poor temporal resolution as a consequence of computing a Fourier Transform over a finite time window. Therefore, the time resolution of the detection signal is only a fraction of the time resolution of the original EEG data acquired. Similarly, machine learning applied to chunked data \cite{kashefi2025epileptic} or pre-detected candidate events\cite{pfammatter2019automated} also presents the same segmentation problem: the beginning and end of an event lack temporal resolution. AugUNet1D avoids this problem altogether by reading the entire, continuous data and treating every sample as a discrete point to label rather than grouping samples into blocks with homogeneous compositions.

Another notable feature of AugUNet1D is its robust performance during periods when other SWD detection methods fail. Time- and frequency-based methods tend to falsely detect events during sleep when the overall EEG amplitude increases. Therefore, methods that rely on changes in EEG amplitude to detect SWDs also tend to detect events during sleep (Figure 8C). Transient, non-pathological EEG oscillations during sleep can increase power in the SWD frequency band (5-7Hz for C3H/HeJ mice ) substantially and are often erroneously detected by thresholding in that frequency band. Overall, false detections during sleep pose a serious problem for time- and frequency-based detection methods. SWDs can happen during sleep\cite{halasz2002spike,sitnikova2021sleep} so excluding sleep altogether is not an adequate solution. AugUNet1D presents a superior alternative by learning the SWD pattern itself rather than its spectral composition or simplified waveform features which can be shared by non-pathological EEG events. 

In conclusion, AugUNet1D is more effective for automatically identifying SWDs in our dataset than all other methods tested here. It outperforms all other traditional machine learning methods, neural networks, and the Twin Peaks algorithm. Furthermore, AugUNet1D is robust, performing well even in conditions where other methods fail (e.g., during slow-wave sleep). We also anticipate that AugUNet1D can be applied more generally to detect a range of EEG signals, including non-SWD EEG events and that AugUNet1D will be valuable for experimenters and clinicians alike; the data presented herein show that trained researchers require nearly one hour to manually label every nine hours of continuous EEG data. AugUNet1D can perform this task much more efficiently and does not require users to actively process data. 


\section*{Data Availability}
Data Availability Statement: Data can be found with this DOI: \url{https://doi.org/10.5281/zenodo.17982389}
Code available here: \url{https://github.com/ssen7/augunet1D}.

\section*{Funding}
This work was supported by the National Institutes of Health under grants R01 NS126594, R01 NS131670, and T32 DK007646.

\bibliography{sample}





\section*{Author contributions statement}
S.Sen and S.K. analyzed results and wrote the manuscript. S.Sen, S.K., D.E.B., and M.B. conceived experiment and analysis approaches and edited the manuscript.  S.Sen developed the primary deep learning architecture. S.Sharma  implemented traditional machine learning approaches for comparison. S.K., S.L., A.M. collected EEG data and manually labeled SWDs.



\section*{Competing interests} 
The authors declare no competing interests related to this work.




\end{document}